\documentclass{article}
\usepackage{spconf,amsmath,graphicx}
\usepackage{booktabs}
\title{Automated Audio Captioning using Transfer Learning and Reconstruction Latent Space Similarity Regularization}
\name{Andrew Koh, Xue Fuzhao, Chng Eng Siong}
\address{Nanyang Technological University}
\begin{document}
%\ninept
%
\maketitle
\begin{abstract}
In this paper, we examine the use of Transfer Learning using Pretrained Audio Neural Networks (PANNs) \cite{kong2020panns}, and propose an architecture that is able to better leverage the acoustic features provided by PANNs for the Automated Audio Captioning Task \cite{drossos_automated_2017}. We also introduce a novel self-supervised objective, Reconstruction Latent Space Similarity Regularization (RLSSR). The RLSSR module supplements the training of the model by minimizing the similarity between the encoder and decoder embedding. The combination of both methods allows us to surpass state of the art results by a significant margin on the Clotho dataset \cite{drossos_clotho_2019} across several metrics and benchmarks.
\end{abstract}
%
% \begin{keywords}
% One, two, three, four, five
% \end{keywords}
%
\section{Introduction}
\label{sec:intro}

Automated Audio Captioning is an emerging research area \cite{drossos_automated_2017} which aims to describe in textual form, the events taking place in the foreground and background of an audio clip. Automated Audio Captioning is the amalgamation of both Audio Processing and Natural Language Processing research, much like Automated Speech Recognition. However, unlike Automated Speech Recognition, Automated Audio Captioning does not require an alignment between the source audio and target text sequence. Furthermore, Automated Audio Captioning requires the model to detect both high level events in the background and the more intricate details that occur in shorter time-frames in the foreground.  %cite this

Convolutional Neural Networks \cite{kong2020panns} are a popular choice of encoder for encoding information from the audio input. Convolutional Neural Networks has done well for tasks such as sound event detection, audio tagging, by being able to extract local features using convolutional filters. This has also led to previous work in Automated Audio Captioning also similarly using a single convolutional neural network as the encoder. However, we argue that this is insufficient for Automated Audio Captioning, as this task requires the model to capture not just local events (e.g a loud honk) that occur in narrower timeframes, but also overarching events (e.g. car engine running in the background) that can span the whole duration of the audio. To combat this, we propose to append the transformer encoder \cite{vaswani2017attention} to the convolutional encoder. The transformer encoder consists of several layers of multi-head self attention and feedforward layers, and many variants of the transformer have broken the state of the art on Natural Language Processing benchmarks \cite{devlin2019bert}. We hypothesise that the combination of both the convolutional encoder and transformer encoder, along with transfer learning, will allow the model to perform much better.

Additionally, we propose a simple self-supervised learning objective, Reconstruction Latent Space Similarity Regularization (RLSSR), to regularize the training of model. As the Automated Audio Captioning is arguably less fine-grained and more abstract in its objective compared to Automated Speech Recognition, we argue that it is insufficient to simply use a cross attention mechanism to bridge information from the audio encoder to the text decoder. RLSSR provides a straightforward way to regularize the training of the decoder by reducing the euclidean distance between the reconstruction of the decoder layer in the latent space and the latent output of the encoder.

Our contribution are as follows:
\begin{enumerate}
    \item We examine the usefulness of the transformer encoder in the encoder, and show that including the transformer encoder is instrumental for the model performance when using transfer Learning.
    \item We introduce the novel Reconstuction Latent Space Similarity Regularization (RLSSR) objective and show that simply using this objective allows a model trained from scratch to achieve competitive state of the art results.
    \item We demonstrate that using both Transfer Learning and RLSSR in tandem consistently beats state of the art results on the Clotho Dataset.
\end{enumerate}

\section{Related Work}
Automated Audio Captioning has been tackled by various authors. 

\subsection{Audio Captioning Datasets} 
\label{audio_cap_dataset}
The Clotho Dataset \cite{drossos_clotho_2019} is by far the dataset most researchers have examined and experimented on. The Clotho Dataset consists of 4981  audio sample of duration from 15 to 30 seconds. Each audio sample are matched to 5 crowdsourced English audio annotations, or captions, that describe the events in that audio. The Clotho Dataset was created to address the critiques of the larger Audiocaps dataset \cite{kim_audiocaps_2019}, which were found to contain biases in the annotation process. There are also niche datasets like the Hospital Scene and Car Scene dataset \cite{xu_audio_2020} which are Mandarin annotated.

In this work, we focus only on the Clotho Dataset, as it is the most popular dataset.

\subsection{Transfer Learning}

Several work has applied transfer learning to Automated Audio Captioning. Fine-tune PreCNN Transformer\cite{chen:2020:dcase:transformer_pretrained_cnn} used a 3 stage pipeline of first pretraining an encoder, then applying transfer learning to the pretrained model by freezing parts of the model for training, and finally finetuning the entire model. 

AT-CNN \cite{xu:2021:ICASSP:02}, the current best performing model on Clotho, follows a similar approach with a 2 pipeline which first pretrains the CNN10 \cite{kong2020panns} / CRNN5 \cite{Dinkel_2020} encoder on an audio tagging task, and then performs transfer learning on the pretrained model on the audio tagging task. Similarly, the Audio Captioning Transformer \cite{mei2021audio} also pretrains the encoder on audio tagging before training on audio tagging.

Other approaches uses specialized embeddings for transfer learning. \cite{koizumi_audio_2020} applied transfer learning to a pretrained GPT2 model, which was pretrained in the text modality. This approach also requires several other prerequisites, such as a BERTScore \cite{zhang2020bertscore} model, and several preprocessing steps for caption retrieval. In another vein, \cite{eren_audio_2021} used a combination of pretrained audio embeddings and subject-verb embeddings to generate captions. 

In this work, we show that applying transfer learning using publicly available pretrained models in conjunction with the transformer encoder allows for significant performance improvements.

% \subsection{Related Approaches} 

\section{Proposed Method}

\begin{figure}[!ht]
  \centering
  \includegraphics[width=\columnwidth]{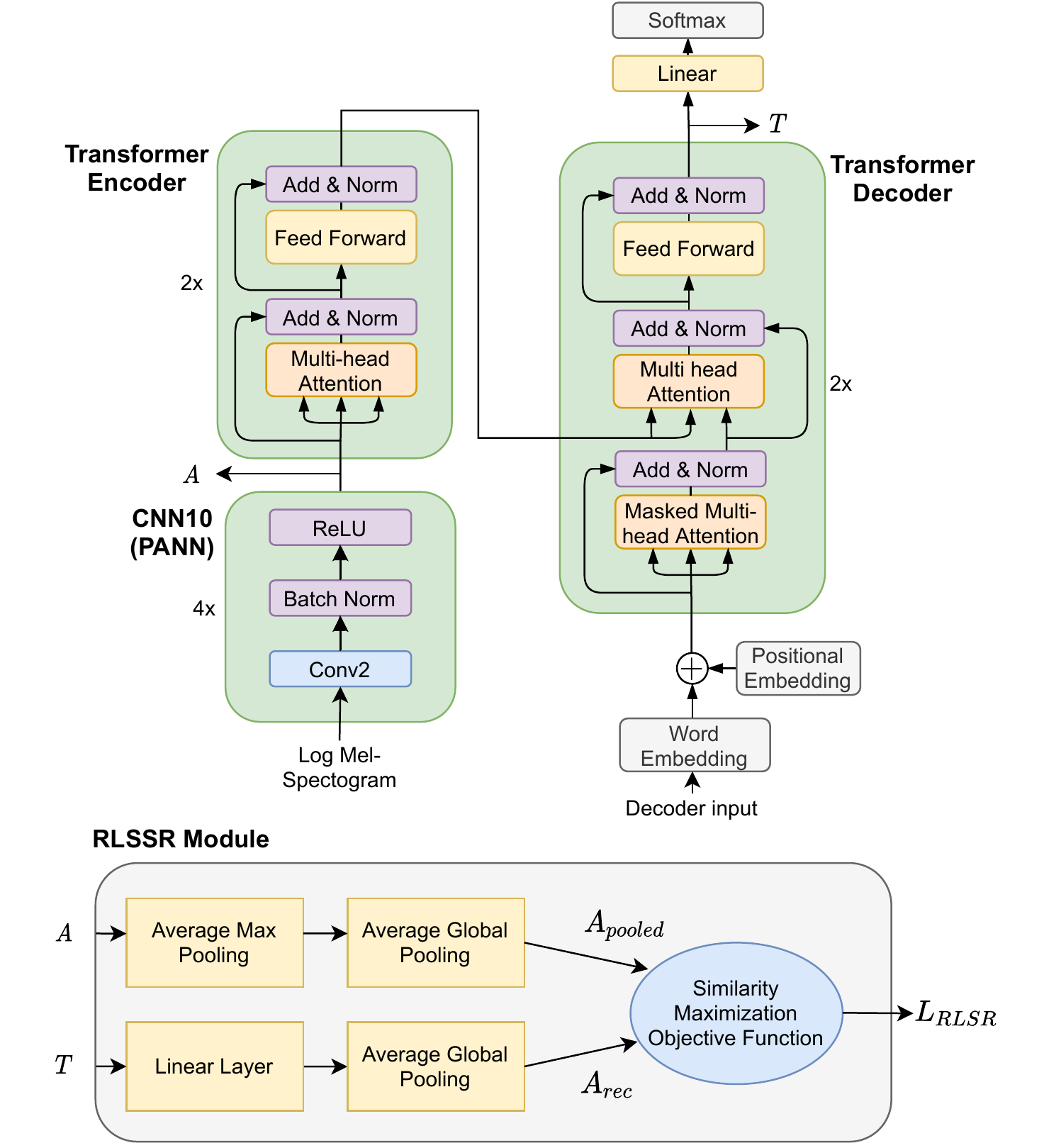}
  \caption{Detailed Overview of the model. The encoder consists of a CNN10 and a Transformer Encoder, and the decoder consists of a single Transformer decoder. The output of the CNN10, $A$, and the output embedding of the transformer decoder, $T$, is passed to the RLSSR module, where the L1/L2 loss is used as objective function in the RLSSR Module.}
  \label{fig:model_overview}
\end{figure}

We use a typical encoder-decoder model in all our experiments. In addition to the encoder-decoder structure, there is an external Reconstruction Latent Space Similarity Regularization (RLSSR) module. A detailed overview of the model is shown in Figure \ref{fig:model_overview}.

\subsection{Encoder}
The encoder consists of a 10 layer CNN (CNN10) and the encoder of the transformer, and takes in log-mel spectrograms as input. CNN10 is adapted from different CNN architectures availables in the Pretrained Audio Neural Networks collection \cite{kong2020panns}. The output of CNN10 is passed to a transformer encoder. The transformer encoder \cite{vaswani2017attention} consists of two layers with a hidden size of 192. We hypothesize that the multihead attention mechanism in the transformer encoder is instrumental in learning higher-level representations from the features provided by the CNN. In our experiments, we show that using the transformer encoder in transfer learning provides a huge boost in performance. Where transfer learning is applied, only the CNN10 encoder is preloaded with pretrained weights from, while the transformer encoder is always trained from scratch.

\subsection{Decoder}
The decoder is a typical Transformer decoder with multihead-attention. It takes a sequence of word tokens, along with the output from the encoder, as decoder input and outputs the next most probable word token. In our experiments, we used a 2 layer Transformer decoder with a hidden size of 192. While we did experiment on decoders of different layer sizes, we found that using 2 layers worked the best.

\subsection{Reconstruction Latent Space Similarity Regularization Module}
% \begin{figure}[ht]
%   \centering
%   \includegraphics[width=\columnwidth]{./images/RLSSR module horizontal.pdf}
%   \caption{The RLSSR module aims to minimize the distance between the reconstruction of the latent space from the decoder and the last layer of the CNN using a distance-based loss function.}
%   \label{fig:RLSSR Module}
% \end{figure}

The aim of the Reconstruction Latent Space Similarity Regularization (RLSSR) module is to maximize the similarity between the latent space of the decoder and the latent space of the encoder. Intuitively, this module has to reconstruct the latent space of the encoder output in the audio domain, $A$, from the latent space of the decoder output in the text domain, $T$. This is done by introducing a very small module of parameters that bridges the decoder output to the encoder.

\begin{equation}
  \begin{array}{l}
    A_{rec} = \text{FFN}(\text{pool}(T)) \\
    A_{pooled}  = \text{pool}(A) \\
    L_{RLSR} = \text{sim}(A_{rec}, A_{pooled}) \\
  \end{array}
 \label{eqn:RLLSR_eqn}
\end{equation}

The RLSSR module takes the output text embedding, $T$ from the last layer of the decoder and the output audio embedding, $A$, from the CNN from the encoder as input. A average max pooling is applied to $A$ to extract the most prominent features to be reconstructed. After which, an average global pooling is applied to standardize the length of the audio features to obtain A\textsubscript{pooled}. The text embeddings $T$ are first passed through a linear layer and then an average global pooling layer to obtain the reconstructed representation from the text, A\textsubscript{rec}. Finally, a similarity metric is used to minimize the distance between A\textsubscript{pooled} and A\textsubscript{rec}. These operations are shown in the Equation \ref{eqn:RLLSR_eqn}. While we did try other similarity metrics such as cosine similarity, Huber loss \cite{girshick2015fast_huber}, we found that the L1 distance, L2 distance worked the best as the similarity metric.

\begin{equation}
\begin{array}{l}
    L = \alpha L_{ce} + \beta L_{RLLSR} \\
 \end{array}
 \label{eqn:loss_combination}
\end{equation}

The model minimizes the sum of the cross entropy loss from the generated caption, along with the loss from the RLLSR module. We tried various combinations of $\alpha$ and $\beta$ and found that weighing both loss values equally worked the best.

\begin{table*}[!ht]
\centering
\resizebox{\textwidth}{!}{%
\begin{tabular}{@{}llllllllll@{}}
\toprule
Model                                    & BLEU\textsubscript{1} & BLEU\textsubscript{2} & BLEU\textsubscript{3} & BLEU\textsubscript{4} & ROUGE\textsubscript{L} & METEOR & CIDEr & SPICE & SPIDEr \\ \midrule
Base (with transformer encoder)                                    & 0.516 & 0.330 & 0.219 & 0.142 & 0.348  & 0.152  & 0.319 & 0.102 & 0.210  \\
\textbf{Base (with transformer encoder) + PANN}            & \textbf{0.542} & \textbf{0.363} & \textbf{0.248} & \textbf{0.163} & \textbf{0.362}  & \textbf{0.161}  & \textbf{0.369} & \textbf{0.108} & \textbf{0.242}  \\
\midrule
Base - transformer enc                   & 0.523 & 0.337 & 0.227 & 0.151 & 0.353  & 0.153  & 0.332 & 0.102 & 0.211  \\ 
Base - transformer enc + PANN            & 0.538 & 0.350 & 0.235 & 0.153 & 0.362  & 0.158  & 0.348 & 0.106 & 0.227  \\
\bottomrule
\end{tabular}
}

\caption{Ablation experiments to determine usefulness of the transformer encoder. Base refers to our default model, consisting of the convolutional encoder, transformer encoder and transformer decoder. PANN refers to loading the pretrained weights for transfer learning.}
\label{tab:trans_enc_ablate}

\vspace{0.01cm}

\resizebox{\textwidth}{!}{%
\begin{tabular}{@{}llllllllll@{}}
\toprule
Model                                    & BLEU\textsubscript{1} & BLEU\textsubscript{2} & BLEU\textsubscript{3} & BLEU\textsubscript{4} & ROUGE\textsubscript{L} & METEOR & CIDEr & SPICE & SPIDEr \\ \midrule
Base                                     & 0.516 & 0.330 & 0.219 & 0.142 & 0.348  & 0.152  & 0.319 & 0.102 & 0.210  \\
Base + L2 loss                 & 0.518 & 0.335 & 0.228 & 0.152 & 0.352 & 0.151 & 0.326 & 0.102 & 0.214    \\ 
Base + L1 loss                 & 0.515 & 0.338 & 0.234 & 0.159 & 0.351 & 0.151 & 0.325 & 0.098 & 0.212    \\ \midrule
Base + PANN                              & 0.542 & 0.363 & 0.248 & 0.163 & 0.362  & 0.161  & 0.369 & 0.108 & 0.242  \\
Base + PANN + L2 loss          & 0.552 & 0.370 & 0.251 & 0.166 & 0.369 & 0.163 & 0.375 & 0.112 & 0.240    \\
\textbf{Base + PANN + L1 loss}          & \textbf{0.551} & \textbf{0.369} & \textbf{0.252} & \textbf{0.168} & \textbf{0.373}  & \textbf{0.165}  & \textbf{0.38}  & \textbf{0.111} & \textbf{0.246}  \\
\bottomrule
\end{tabular}
}
\caption{Results of using the RLSSR module. L1 and L2 loss refers to the distance metric used to optimize the RLSSR module. PANN refers to applying transfer learning using the pretrained weights from the pretrained audio neural network.}
\label{tab:RLSSR results}

\vspace{0.01cm}

\resizebox{\textwidth}{!}{%
\begin{tabular}{@{}llllllllll@{}}
\toprule
Model                                    & BLEU\textsubscript{1} & BLEU\textsubscript{2} & BLEU\textsubscript{3} & BLEU\textsubscript{4} & ROUGE\textsubscript{L} & METEOR & CIDEr & SPICE & SPIDEr \\ \midrule
Baseline \cite{drossos_clotho_2019}                                 & 0.389 & 0.136 & 0.055 & 0.015 & 0.262 & 0.084 & 0.074 & 0.033 & 0.054    \\
Fine-tune PreCNN Transformer \cite{chen:2020:dcase:transformer_pretrained_cnn}             & 0.534 & 0.343 & 0.230 & 0.151 & 0.356 & 0.160 & 0.346 & 0.108 & 0.227    \\
AT-CNN10 \cite{xu:2021:ICASSP:02}                                 & 0.556 & 0.363 & 0.242 & 0.159 & 0.368 & \textbf{0.169} & 0.377 & \textbf{0.115} & - \\

\textbf{Base + PANN + L1 loss}          & \textbf{0.551} & \textbf{0.369} & \textbf{0.252} & \textbf{0.168} & \textbf{0.373}  & 0.165  & \textbf{0.380}  & 0.111 & \textbf{0.246}  \\
\bottomrule
\end{tabular}
}
\caption{Comparison with other state of the art. Our model consistently beats the previous state of the art on the BLEU\textsubscript{n}, ROUGE\textsubscript{L}, and CIDEr scores.}
\label{tab:comparison_sota}
\end{table*}

\section{Experimental Details}
\subsection{Data}
As mentioned in Section \ref{audio_cap_dataset}, the experimental dataset in use is the Clotho dataset. We try to follow as closely as possible the experimental settings of previous work. These are the specific details. The raw audio files are first preprocessed into log mel-spectrograms. We use 64 Mel-bands, sampling rate of 44100, FFT window length of 1024, and a hop size of 512.

\subsection{Training and Evaluation}

These are our training hyperparameters and settings. We use a batch size of 64 with a gradient accumulation steps of 4 for 200 epochs with early stopping. Learning rate is set to $3 \times 10^{-4}$ and SpecAugmentation\cite{specaug} is applied to all log mel-spectrogram inputs as an data augmentation tactic. We do not apply label smoothing. The weights of the cross entropy loss and the similarity loss from the RLSSR module is weighted and optimized equally. Where transfer learning is applied, we used the weights of the CNN10 model\footnote{This pretrained model is available publicly for download at https://github.com\/qiuqiangkong\/audioset\_tagging\_cnn} which scored a mAP of 0.380 on Audioset. 

For inference, we perform beam search with beam size of 4 for decoding. As the Clotho dataset has 5 reference captions for each audio, we follow the COCO image captioning evaluation process \cite{coco_dataset} to evaluate the generated caption. The BLEU\textsubscript{n} scores \cite{bleu_cite} calculates the degree of n-gram overlap between the generated and reference caption. ROUGE\textsubscript{L} \cite{lin-2004-rouge} measures the longest common sequence and gives a score base on the similarity. METEOR \cite{denkowski:lavie:meteor-wmt:2014} is said to better correlate with human annotators and uses a harmonic mean of unigram precision and recall to score the generated sequence.
CIDEr \cite{cider_cite} also measures similarity using the average cosine similarity between the candidate sentence and the reference sentences. Like METEOR, SPICE \cite{anderson2016spice} is also said to correlate with human judgement more by using semantic propositional content and is less reliant on n-gram overlaps. Finally, SPIDEr \cite{spider} is a linear combination of both SPICE and CIDEr. We use all the aforementioned metrics for evaluation.

% Our code is available at <link here>
\begin{table*}[h!]
\centering
\resizebox{\textwidth}{!}{%
\begin{tabular}{|p{0.55\linewidth}|p{0.25\linewidth}|p{0.25\linewidth}|}
% \begin{tabular}{|l|l|l|}
\hline
\textbf{Reference}                                                                                        & \textbf{base-enc}                                       & \textbf{base+PANN+L1}                                                 \\ \hline
water flowing at a constant rate then slows down to a drip                                       & water is running into a sink or tub            & water is running at a constant rate in a sink                \\ \hline
a person is arranging objects on a table and pouring water into a glass                          & someone is washing dishes and dishes in a sink & water is being poured into a container while a person speaks \\ \hline
a man slides the file across the wood tapping it a few times at the end & a door is opened and closed several times      & a hard object is being dragged across a hard surface         \\ \hline
\end{tabular}
}
\caption{Table containing the generated examples of our models, along with the reference captions.}
\label{tab:examples_generated}
\end{table*}

\section{Experimental Results}
\subsection{Usefulness of the Transformer Encoder}

We compare the effectiveness of including the transformer encoder in the encoder. The results are shown in Table \ref{tab:trans_enc_ablate}. Experiments show that using pretrained weights along with the transformer encoder outperformed all the other models by a significant margin. On the other hand, not including the transformer encoder along with pretrained weights resulted in only marginal improvements compared to not using pretrained weights. It is also interesting to note that, without pretrained weights, not using the transformer encoder boosted the performance. We believe this is because the size of the Clotho dataset is insufficient to meaningfully train a transformer encoder from scratch. However, using pretrained weights in the convolutional encoder provided the initialization required for the transformer encoder to learn much better.

\subsection{Effect of the Reconstruction Latent Space Similarity Regularization Module}

We examine the impact of the RLSSR Module. Results are shown in Table \ref{tab:RLSSR results}. From our experiments, using the L1 loss outperforms the L2 loss consistently. However, using either the L1 loss or L2 loss in RLSSR module outperforms the models which did not use the RLSSR module. Models which did not use transfer learning but used the RLSSR module outperformed the base model by a significant margin and even performed competitively to state of the art models using transfer learning. When using transfer learning, the performance of our models increased even further. This attests to the RLSSR module being an useful objective to optimize to effectively regularize the model.

\subsection{Comparison with previous work}
The comparison with other work is shown in Table \ref{tab:comparison_sota}. Our best model beats the best performing model, AT-CNN10 \cite{xu:2021:ICASSP:02}, on all metrics except the METEOR and SPICE metric. In addition to being self-supervised and simple to train, our approach makes use of publicly available pretrained models, while AT-CNN10 involves a two stage transfer learning process trained from scratch. We have also included some examples of our generated sequences at Table \ref{tab:examples_generated}.

\section{Conclusion}
This work introduces the benefit of the transformer encoder when applying transfer learning. We have also proposed the self-supervised Reconstruction Latent Space Similarity Regularization objective, which acts as regularization during training. We have shown that combining both RLSSR and transfer learning allows us to beat state of the art on the Clotho dataset by a significant margin across many metrics.

\bibliographystyle{IEEEbib}
\bibliography{refs}

\end{document}